\def\year{2019}\relax
\documentclass[letterpaper]{article} 
\usepackage{aaai19}  
\usepackage{times}  
\usepackage{helvet}  
\usepackage{courier}  
\usepackage{url}  
\usepackage{graphicx}  
\usepackage{amsmath}
\usepackage{amsthm}
\begin{document}
%
\title{{Adaptive Region Embedding for Text Classification}}
\author{Liuyu Xiang$^1$, Xiaoming Jin$^1$, Lan Yi$^2$, Guiguang Ding$^1$\\
$^1$School of Software, Tsinghua University, Beijing, China\\
$^2$Department of DevNet, Cisco Systems\\
xiangly17@mails.tsinghua.edu.cn, xmjin@tsinghua.edu.cn,  \\
layi@cisco.com, dinggg@tsinghua.edu.cn \\
}
\maketitle
\begin{abstract}

  Deep learning models such as convolutional neural networks and recurrent networks are widely applied in text classification. In spite of their great success, most deep learning models neglect the importance of modeling context information, which is crucial to understanding texts. In this work, we propose the \emph{Adaptive Region Embedding} to learn context representation to improve text classification. Specifically, a \emph{meta-network} is learned to generate a context matrix for each region, and each word interacts with its corresponding context matrix to produce the regional representation for further classification. Compared to previous models that are designed to capture context information, our model contains less parameters and is more flexible. We extensively evaluate our method on 8 benchmark datasets for text classification. The experimental results prove that our method achieves state-of-the-art performances and effectively avoids word ambiguity.

\end{abstract}

\section{Introduction}

Text classification is a fundamental task in many NLP applications such as web searching, information retrieval and sentiment analysis. 
One key challenge in text classification is to understand the \emph{compositionality} in text sequences for which the modeling of regional relationships between words and their neighbours is crucial. Traditional methods usually exploit n-grams as composition function, which proves to be effective by directly gathering adjacent words. With the emerging development of deep learning, a variety of deep models are recently used for text classification. The most commonly used models are Recurrent Neural Networks (RNNs) \cite{tang2015document,yang2016hierarchical,zhou2016text} and Convolutional Neural Networks (CNNs) \cite{kim2014convolutional,zhang2015character}. 
RNN and its variants LSTM/GRU model the regional relationships by encoding the previous words into its hidden unit while CNN captures the compositional structure by layers of convolution and pooling. In spite of their success on various NLP tasks, RNNs and CNNs are only generic models for sequences or images. They may neglect the intrinsic semantic relations in text sequences and often require large computational cost and memory space. 

Subsequently, several simple and effective models have been proposed to improve the text classification by learning a specially designed high-level representation that contains the necessary regional context information. Here we loosely refer to these learned  compositional representations as `\textbf{Region Embedding}'. 
Among these methods, \cite{johnson2015semi} learn three types of region embeddings and feed them to a CNN as extra inputs. The region embeddings help to improve the error rates by a significant margin compared to vanilla CNN. \cite{qiao2018a} propose a two-layer architecture, and introduce a matrix called `Local Context Unit' to produce region embeddings. They achieve comparable or even superior performances to a 29-layer CNN \cite{conneau2017very}, which is the current state-of-the-art method.  

Although \cite{qiao2018a}'s region embedding method is able to effectively capture rich context information, it still has the following limitations. 
First, the region embedding is generated by word embedding and Local Context Unit, where both of them are only and uniquely determined by word indices in the vocabulary. Such embedding method lacks the capability to capture the semantic ambiguity since each word has the same unique region embedding even under different contexts.

Second, its representation capability is limited by the fact that the Local Context Unit is generated within just a fixed-length region. This leads to the negligence of richer dependencies outside the region.  Third, the embedding look-up tensor that generates Local Context Unit is of size $E \in R^{v \times h}$, where $v, h$ are sizes of vocabulary and embedding dimension, respectively. For dataset such as Yahoo Answer with \textasciitilde{}300k words in the vocabulary, i.e. $v \approx 300k$, the storage of embedding matrix $E$ requires large memory resources.

To address the limitations of \cite{qiao2018a}'s method, we need to design a light-weighted and flexible network component which is able to take the whole sequence into account, and produce effective region embedding adaptive to different contexts. 

Inspired by the recent work of dynamic parameter generation \cite{bertinetto2016learning,jia2016dynamic,ha2016hypernetworks}, we propose a novel region embedding method in this paper. First we use a \emph{meta-network} \cite{ha2016hypernetworks} to distill meta knowledge from the whole sequence and generate the context matrix (or context unit) for each small region of texts. Then the region embedding is produced by both the context unit and the word embeddings in the corresponding region. The meta-network we use here can be of \textbf{any form of fully differentiable neural network} that can be jointly trained end-to-end. Our meta-network only requires a tiny amount of parameters, leading to much smaller memory cost. 
We call our context unit generated by the meta-network \emph{Adaptive Context Unit} to distinguish from \emph{Local Context Unit} \cite{qiao2018a}. For simplicity, we refer to Local Context Unit and Adaptive Context Unit as \textbf{LCU} and \textbf{ACU}, and Local Region Embedding and Adaptive Region Embedding as \textbf{LRE} and \textbf{ARE} respectively in the following discussion. 

We compare our method with \cite{qiao2018a}'s and previous state-of-the-art methods on eight benchmark text classification datasets. The experimental results demonstrate that our method is able to achieve state-of-the-art results with much less parameters than \cite{qiao2018a}'s method. 

Apart from the experimental superiority, we also try to explore the mechanisms behind. We study the differences and relationships between our method, \cite{qiao2018a}'s method and the normal convolution, and generalize them to instance-level, word-level, dataset-level respectively from the perspective of \emph{Generalized Text Filtering}.
This generalization partly explains why our proposed method is more flexible and performs better.

To sum up, the main purpose of this work is to learn a more flexible and compact region embedding. Our main contributions are threefold:
\begin{itemize}
    \item We propose a new region embedding method where richer context information is acquired through a meta-network.
    \item Our method achieves state-of-the-art performances on several benchmark datasets with a small parameter space. We also demonstrate that our model is able to avoid word ambiguity. 
    \item We generalize convolution, \cite{qiao2018a} and our method under the same framework, which reveals the mechanisms accounting for our method's superiority.
\end{itemize}

\section{Related Work}

\subsection{Text Classification}
Text classification has long been studied in Natural Language Processing. Before the deep learning era, traditional methods usually consist of high-dimensional features followed by classifiers such as SVM and logistic regression \cite{joachims1999transductive,fan2008liblinear,mccallum1998comparison}. Bag-of-words (BoW) and n-grams are the commonly used features. However, BoW usually suffers from the ignorance of word order, whereas n-grams usually suffers from the notorious \emph{curse of dimensionality}. Besides, traditional methods \cite{joachims1999transductive,fan2008liblinear,forman2003extensive,li2012multi} usually rely heavily on hand-crafted features or graphical models, which are often laborious and time-consuming. 

In recent years, the emergence of distributed word representations \cite{Mikolov2013DistributedRO} and deep learning has enabled automatic feature learning and end-to-end training, providing superior performances over the traditional methods on various NLP tasks. 
Most deep learning models that are used for text classification are based on RNN \cite{sundermeyer2012lstm,yang2016hierarchical,yogatama2017generative}, or CNN \cite{kim2014convolutional,johnson2015effective,zhang2015character,conneau2017very}. While these deep models are equipped with more complex composition function to aggregate the separate word embeddings into a single semantic vector, they usually require large parameter space and computational cost.

Apart from these deep models, there are also other simple and effective models adopted as composition function for text classification. 
\cite{johnson2015semi} design an extra network that represents the context of each word by learning to predict the neighbours of the word unsupervisedly, and use that extra network to produce context-related features as additional inputs. FastText \cite{joulin2017bag} utilizes the averages of distributed word embeddings as inputs to a hierarchical softmax.
\cite{Shen2018BaselineNM} apply hierarchical pooling to word embeddings and achieve comparable performances to the CNN/LSTM based methods. \cite{wang_id_2018_ACL}'s method leverages the label information as attention map, and learns a joint embedding of words and labels.
\cite{qiao2018a} choose to represent the context information with the LCU, and produce its region embedding by the projecting the word embedding of each region onto the context unit. 
Our work is also similar to another contemporary work \cite{shen2018learning} where both methods utilize the meta-network structure. However, our work differs from \cite{shen2018learning} in both the particular meta-network used and the whole architecture. Moreover, our proposed \emph{Generalized Text Filtering} perspective provides a generalization of three levels of text filtering, and will generalize both \cite{shen2018learning} and \cite{qiao2018a}'s methods.

\begin{figure*}[!h]
\centering
\includegraphics[width=0.9\textwidth]{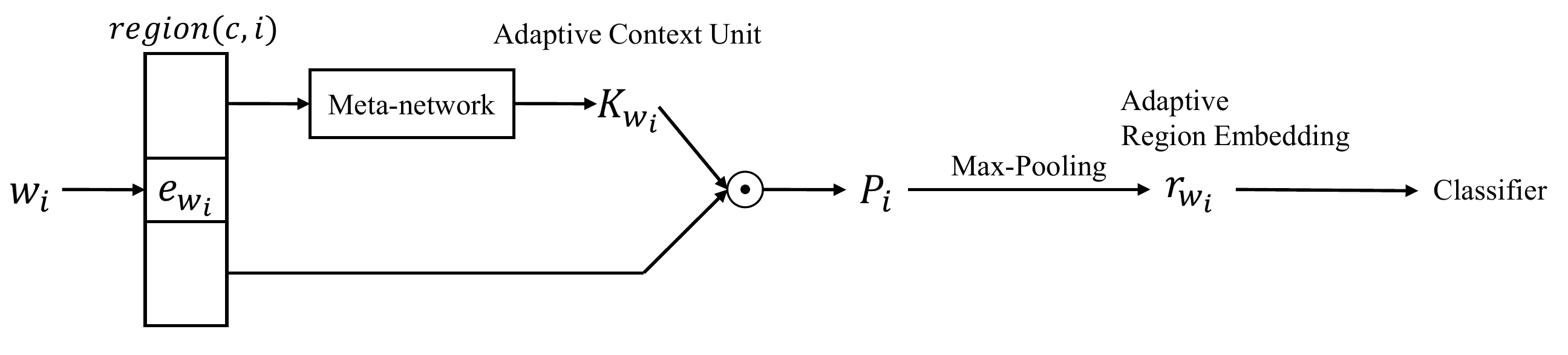}
\caption{Overall Architecture of Adaptive Region Embedding}
\end{figure*}

\subsection{Dynamical Weight Generation}
Another related research area is the \emph{Dynamical Weight Generation}, which refers to a special kind of network architecture, where the parameters of the basic network are generated by a higher-level network, which is called meta-network. This structure  \cite{jia2016dynamic,ha2016hypernetworks,bertinetto2016learning,chen2018meta} usually involves two kinds of parameters: \emph{meta-network parameters} and \emph{dynamically generated parameters}. The \emph{meta-network parameters} are learned through gradient descent, whereas the \emph{dynamically generated parameters} are generated by the \emph{meta-network parameters} with each instance as input, thus they are \textbf{instance-specific}. Besides, the parameter space of the meta-network is usually small, so that the whole structure is practical in large-scale applications.

In the literature, \cite{jia2016dynamic} mainly focus on generating dynamic convolutional filters conditioned on the input. \cite{ha2016hypernetworks} further extend the meta-network structure to both CNN and RNN-based structures. They also put forward factorization techniques to make the meta-network scalable and memory-efficient. Then meta-network structures are applied to several other applications. \cite{bertinetto2016learning} model one-shot learning as dynamic parameter prediction, and use a siamese-like structure to enable one-shot learning. \cite{chen2018meta} adopt an LSTM-based meta-network structure for multi-task sequence modeling, where the meta-network module is shared between tasks, and it controls the parameters of task-specific private layers. 

The insight behind using the meta-network is that the parameters/weights that multiply with the input features are reduced from \textbf{dataset-level} to \textbf{instance-level} so that they are more adaptive. Take CNN for example, normal CNN kernels or filters are learned and updated through the whole dataset by gradient descent, and do not adapt to any particular instance, where in meta-network structure the \emph{dynamically generated weights} can adapt to any specific instance, so that the whole network is able to capture richer information.

\section{Proposed Method}

\subsection{Overview}
The proposed model architecture is shown in Figure 1. Its core component is the meta-network-generated ACU. Given a region centered at position $i$ with radius $c$ (which we denote as $region(c,i)$), our model first generates the ACU with a meta-network, and then applies the 
ACU back to the region itself to produce the ARE for final classification. The architecture we propose here is similar to that in \cite{qiao2018a}, where both models achieve state-of-the-art text classification performances with a context unit as the core component. However, the conversion from LCU to ACU brings several benefits, including higher classification accuracies, smaller memory consumption and word ambiguity avoidance. We will also discuss further why ACU will result in the improvements mentioned above from a generalized filtering perspective.

Since our main contribution lies in the ACU, as a more flexible version of LCU, we will first briefly introduce the word preprocessing procedure and \cite{qiao2018a}'s methods, then put our main focus on how ACU is generated.

\subsection{Preliminary}
\paragraph{Word-level Preprocessing} 
We implement the whole network at word-level, that is, we use a one-hot vector $w_i$ to represent each word at position $i$ and then transform each word to an $h$-dimension continuous word embedding $e_i$, using an embedding look-up matrix $\mathbf{V} \in R^{h \times v}$, where $h$ and $v$ are the embedding size and vocabulary size respectively. We \textbf{do not} use any pre-trained word embedding as initialization in our model. 

\paragraph{Local Context Unit}
In \cite{qiao2018a}, the authors introduce LCU to interact with the words in a given region to generate the region embedding LRE. The core idea is that the LCU of a word is determined by retrieving a look-up tensor U with the given word index. To be specific, if we use $h, c, v$ to denote the embedding size, region radius and vocabulary size respectively, the LCU for a given word $w_i$ is represented as a matrix $\mathbf{K}_{w_i} \in R^{h\times (2c+1)}$, which contains all the information from $w_i$'s `viewpoint' and is calculated by looking up $w_i$'s index in the look-up tensor $\mathbf{U} \in R^{h \times (2c+1)\times v}$ . Consequently the whole LCU for a given sequence is $\mathbf{K} \in R^{h\times (2c+1) \times L} $, where $L$ is the length of the sequence.

The interaction between a word and its corresponding LCU is to project the word embedding by the LCU $\mathbf{K}_{w_i}$: 
\[ \mathbf{p}_{w_i+t}^i = \mathbf{K}_{w_i,t} \odot e_{w_i+t}, \quad -c \leq t \leq c\] 
Where $e_{w_i+t}$ is the word embedding of word $w_{i+t}$, and $\odot$ denotes element-wise multiplication. By doing projection, the region embedding $\mathbf{r}_{(i,c)}$ for  $region(c,i)$ is built by max-pooling the projected embedding $\mathbf{p}_{w_{i+t}}^i$. The authors also propose two types of pooling schemes to produce the LRE, which we will
not discuss in detail in this paper.

\subsection{Adaptive Region Embedding}
In this section, we deliberate on how ACU is constructed, how it interacts with the words in the region to produce ARE. 

Our model takes a matrix of word embeddings $\mathbf{E} = [e_0, e_1, ..., e_L] \in R^{h \times L}$ as input, where $e_i$ is the word embedding of $w_i$ in a given sequence. Then the ACU is generated by a fully-differentiable meta-network, which takes the original sequence embedding as input, and output the parameters of a base learner. To be more specific, it takes the input of all the word embeddings $E$ in the sequence, and output the ACU: $\mathbf{K} \in R^{h \times (2c+1) \times L}$, where each element $\mathbf{K}_{w_i} \in R^{h \times (2c+1)}$ can be seen as filters or projection matrix for $region(i,c)$ centered at word $w_i$.

In this work, we choose our meta-network to be a one-layer convolutional neural network, such that,

\[ \mathbf{K} = Bn( Conv(\mathbf{E}))\]

where $Conv$ stands for 1-d convolution, and $Bn$ stands for batch normalization layer \cite{ioffe2015batch} for the purpose of reducing covariate shift. The 1-d convolution layer has $h$ input channels, and $h\times(2c+1)$ output channels. Thus the output of the convolution $\mathbf{K} \in R^{h \times (2c+1) \times L}$ meets the requirement to be the context unit. It is worth noting that although the region size is fixed, the `receptive field' of ACU is not limited to $region(i,c)$ since $\mathbf{K}$ is produced by the meta-network which takes the whole sequence as input. Therefore, unlike LCU which is determined only by the region itself, ACU is more flexible and adaptive.

The above calculation outputs the ACU, i.e. $\mathbf{K}$, which is generated at instance-level as we explained in Section 2.2. Then we project the word embedding $\mathbf{E}$ into the region embedding space using the ACU, 
\[\mathbf{p}_{i} = \mathbf{K}_{w_i} \odot \mathbf{E}_{i-c:i+c}\]
where $\mathbf{E}_{i-c:i+c} = [e_{i-c}, ..., e_{i+c}]$, and $\mathbf{p}_{i} \in R^{h\times (2c+1)}$ represents the projected embedding for $region(i,c)$. Each column $\mathbf{p}_{i,w_t}$ represents the projected embedding at position $i$ filtered by $w_t$'s information.

Finally we pool the projected embedding within each $region(i,c)$:
\[ \mathbf{r}_i=g(\mathbf{p}_{i, w_{i-c:i+c}})\] 
where $\mathbf{r}_i \in R^{h}$ represents the ARE at position $i$, and $g$ is the pooling function. Here we choose $g=max()$ which stands for max-pooling along the second dimension of $\mathbf{p}_i$.
Finally, the ARE for the whole sequence $\mathbf{r}$ is calculated by summing region embeddings at all positions $\mathbf{r} = \sum_{i} \mathbf{r}_i$.

After obtaining the ARE $\mathbf{r}$, we feed it into a fully-connected layer followed by a softmax layer, 
\[ \mathbf{y} = Softmax(\mathbf{Wr + b}) \]
Where $\mathbf{W, b}$ are learnable parameters in the fully-connected layer. We choose cross entropy as our loss function.

\section{Generalizing Region Embedding and Convolution}

\begin{figure}[!h]
\centering
\includegraphics[width=0.5\textwidth]{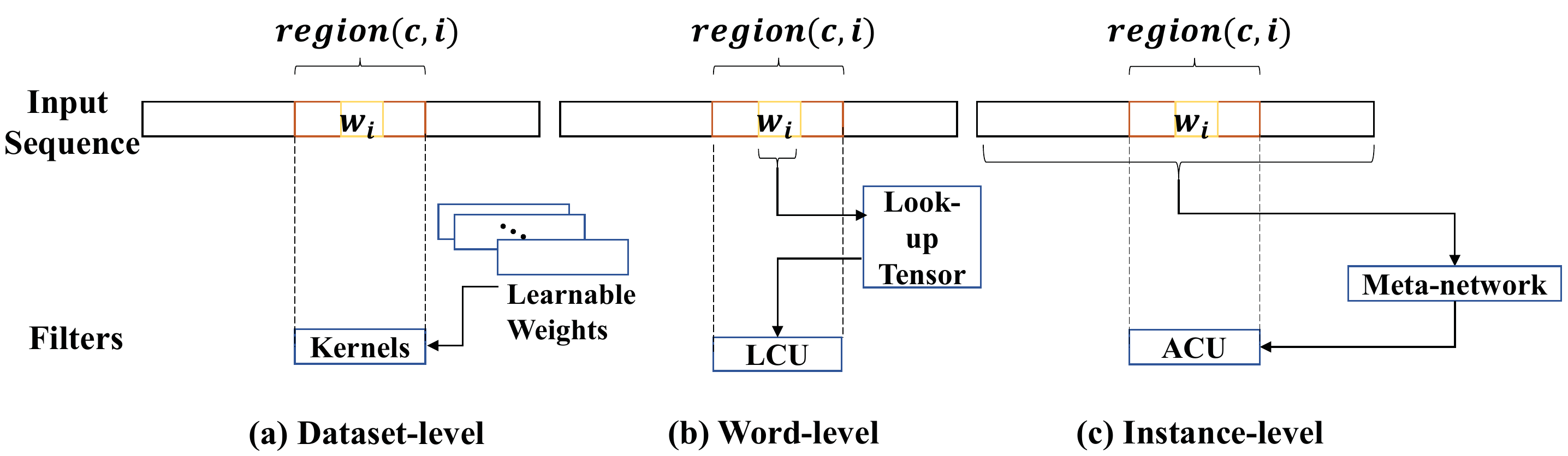}
\caption{Illustration of three levels of filtering.}
\end{figure}

In this section, we will analyze
\cite{qiao2018a}'s LRE method, our ARE method, and convolution in CNN from the aspect of generalized framework \emph{Generalized Text Filtering}, which will partly account for the superiority of our proposed method.

Assume $L$, $L^{'}$, $h$, $r$ denote the input sequence length, output sequence length, word embedding size and window size, respectively, we define \emph{Generalized Text Filtering} as below:

\theoremstyle{definition}
\newtheorem{definition}{Definition}[]
\begin{definition}{\emph{Generalized Text Filtering} $f$ takes the input sequence $\mathbf{x}_{0:L-1} = [\mathbf{x}_0, \mathbf{x}_1, ..., \mathbf{x}_{L-1}] \in R^{h\times L}$, and output the filtered sequence $\mathbf{y} \in R^{L'}$ with one filter $\mathbf{W} \in R^{r\times h}$}
\[ \mathbf{y} = f(\mathbf{x}_{0:L-1}) = [y_0, y_1, ..., y_{L'}]\]
\[ y_i = g(\mathbf{w}_0^{T}\mathbf{x}_i,\mathbf{w}_1^{T}\mathbf{x}_{i+1},...,\mathbf{w}_{r-1}^{T}\mathbf{x}_{i+r-1}) , \quad i \in [0, L']\]

where each $\mathbf{w}_k \in R^{h}$ is the filter at position $k$, $k \in [0, r-1]$, function $g$ is the pooling function that can be either $g=max()$ or $g=sum()$. \end{definition}

When $g=sum()$, the \emph{Generalized Text Filtering} equals to normal convolution. In \cite{qiao2018a} and our method, $g=max()$. 

Regarding the definition of \emph{Generalized Text Filtering} , convolutional kernels, LCU and ACU can all be regarded as special instances of filters in \emph{Generalized Text Filtering}. The process of generating region embedding or convolved features can be regarded as the process of filtering the text with the corresponding filters $W$. 

Moreover, the relationship between these methods is that they correspond to three different levels of filtering. The convolution uses the same set of filters that are shared by all the instances in the dataset, thus corresponding to \textbf{dataset-level filtering}. \cite{qiao2018a} use filters (LCU) uniquely determined by the words in the region center, and they are generated from a look-up tensor by the word's index in the vocabulary, thus corresponding to \textbf{word-level filtering}. 
Our ACU filters however, are generated by feeding the whole sequence of an instance into a meta-network, which we assume will acquire general knowledge during joint training, thus corresponding to \textbf{instance-level filtering}.
The generated ACU filters are expected to be capable of not only exploiting global information, but also adapting to each instance and capturing detailed compositionality. One of the benefits of adopting ACU is the avoidance of word ambiguity. Since LCU filters for each word are uniquely stored in the look-up tensor $\mathbf{U}$, they remain the same even under different contexts where our ACU filters solve such problem thanks to the flexibility brought by the meta-network. An illustration of the three types of filtering can be found in Figure 2.

Our key insights are as follows: 
The usage of a jointly trained meta-network acquires meta-knowledge among texts, and the learned general knowledge is transferred to the generated adaptive filters ACU so that it can capture local compositionality more effectively than convolution, while also being more adaptive to specific context than LCU.

\begin{table*}

  \label{table}
  \centering
  \begin{tabular}{lllllllll}
    \hline
    Dataset & Yelp P. & Yelp F. & Amaz. P. & Amaz. F. & AG & Sogou & Yah. A. & DBP

    \\ [0.5ex] 
    \hline
    \hline
    BoW & 92.2 & 58.0 & 90.4 & 54.6 & 88.8 & 92.9 & 68.9 & 96.6 \\ [0.5ex] 
    ngrams & 95.6 & 56.3 & 92.0 & 54.3 & 92.0 & 97.1 & 68.5 & 98.6\\[0.5ex] 
    ngrams TFIDF & 95.4 & 54.8 & 91.5 & 52.4 & 92.4 & 97.2 & 68.5 & 98.7 \\[0.5ex] 
    \hline 
    char-CNN \cite{zhang2015character} & 94.7 & 62.0 & 94.5 & 59.6 & 87.2 & 95.1 & 71.2 & 98.3\\ [0.5ex] 
    
    VDCNN \cite{conneau2017very} & 95.7 & 64.7 & 95.7 &
    \textbf{63.0} & 91.3 & 96.8 & 73.4 & 98.7\\ [0.5ex] 
    
    FastText \cite{joulin2017bag} & 95.7 & 63.9 & 94.6 & 60.2 & 92.5 & 96.8 & 72.3 & 98.6\\[0.5ex] 

    SWEM \cite{Shen2018BaselineNM} & 93.8 & 61.1 & - & - & 92.2 & - & 73.5 & 98.4 \\[0.5ex] 
    LEAM \cite{wang_id_2018_ACL} & 95.3 & 64.1 & - & - & 92.5 & - & \textbf{77.4} &
    99.0\\[0.5ex] 
    LRE \cite{qiao2018a} & 96.4 & 64.9 & 95.3 & 60.9 & 92.8 & \textbf{97.6} & 73.7 & 98.9\\[0.5ex] 
    \hline

    ARE (Ours) & \textbf{96.6} & \textbf{65.9} &\textbf{95.9} & 62.6 & \textbf{93.1} & 97.5 & 74.9 & \textbf{99.1} \\
    \hline
  \end{tabular}
\caption{Classification accuracies on 8 benchmarks. Traditional methods' result come from \cite{zhang2015character}, all deep learning baselines' results come from their original paper.}
\end{table*}

\section{Evaluation}

\begin{table}

  \label{sample-table}
  \centering
  
  \begin{tabular}{lllll}
    \hline
    Dataset & Train Size & Test Size & Class & Vocab Size\\ [0.5ex] 
    \hline
    AG & 120,000 & 7,600 & 4 & 42783 \\ [0.5ex] 
    Sogou & 450,000 & 60,000 & 5 & 99394\\[0.5ex] 
    DBP. & 560,000 & 70,000 & 14 & 227863\\[0.5ex] 
    Yelp.P & 560,000 & 38,000 & 2 & 115298\\[0.5ex] 
    Yelp.F & 650,000 & 50,000 & 5 & 124273\\[0.5ex] 
    Yah.A. & 1,400,000 & 60,000 & 10 & 361926\\[0.5ex] 
    Amaz.P & 3,600,000 & 400,000 & 2 & 394385\\ [0.5ex] 
    Amaz.F & 3,000,000 & 650,000 & 5 & 356312\\  [0.5ex] 
    
    \hline
  \end{tabular}
  \caption{Detailed information of datasets}
\end{table}

\subsection{Datasets and Tasks}

We report results on 8 benchmark datasets for large-scale text classification. These datasets are from \cite{zhang2015character} and the tasks involve topic classification, sentiment analysis, and ontology extraction. The details of the dataset can be found in Table 2.

\begin{table*}[h]
  \label{sample-table}
  \centering
  \begin{tabular}{cp{2.5cm}ccccc}
    \hline
    Params & \cite{qiao2018a}'s Total & Ours Total & LCU Only {} & ACU Only {} \\ [0.5ex] 
    \hline
    \hline
    AG's news & 43,810,308 & 16,268,804 & 38,333,568 & 5,315,328\\ [0.5ex] 
    Sogou News & 101,780,101 & 30,761,477 & 89,057,024 & $\sim$ \\[0.5ex] 
    DBPedia & 233,333,518 & 63,651,854  & 204,165,248 & $\sim$ \\[0.5ex] 
    Yelp Review Polarity & 118,065,410 & 34,832,130  & 103,307,008 & $\sim$\\[0.5ex] 
    Yelp Review Full & 127,256,197 & 37,130,501 &111,348,608 & $\sim$ \\[0.5ex] 
    Yahoo! Answers & 370,613,514 & 97,970,954 & 324,285,696 &$\sim$ \\[0.5ex] 
    Amazon Review Polarity & 403,850,498 & 106,278,402 & 353,368,960 & $\sim$ \\ [0.5ex] 
    Amazon Review Full & 364,864,133 & 96,532,485  & 319,255,552 & $\sim$ \\  [0.5ex] 
    
    \hline
  \end{tabular}
  \caption{Comparison of the number of parameters. `$\sim$' in the last column indicates that the number of parameters remains the same on different datasets. It is worth noting that these parameters correspond to the models that achieve the best performances where for \cite{qiao2018a} region size and embedding size are 7 and 128, and for ours are 9 and 256 respectively.}
\end{table*}

\subsection{Baselines}
Our baselines include traditional methods and deep learning methods.
For traditional methods, BoW, ngrams, ngrams and its TF-IDF is used as hand-crafted features, and logistic regression is used as the classifier.
For deep learning methods, char-CNN \cite{zhang2015character} and VDCNN \cite{conneau2017very} are both deep CNN-based models. FastText \cite{joulin2017bag}, SWEM \cite{Shen2018BaselineNM}, LEAM \cite{wang_id_2018_ACL} and LRE \cite{qiao2018a} are other neural network based methods. Among them, VDCNN \cite{conneau2017very} which is as deep as 29 layers, LEAM \cite{wang_id_2018_ACL} and LRE \cite{qiao2018a} are the previous state-of-the-art methods.

\subsection{Implementation Details}
\paragraph{Input}
We use the same data preprocessing procedure as \cite{qiao2018a}, that we convert all the words into lower case, and tokenize them using Standford tokenizer. The words that only appeared once are removed out of the vocabulary and we pad each document with length $c$ at the start and the end for filtering. Then each word is represented as a one-hot vector by filling $1$ at its index in the vocabulary.
\paragraph{Hyperparameters} We tune the region size $(2c+1)$ to be 9, embedding size to be 256. We also discuss the impact of these hyperparameters in the following paragraph.
\paragraph{Training} Since the meta-network is fully differentiable, we train them together with the whole model with the same optimizer and learning rate. We choose the batch size to be 16 and the learning rate to be $1 \times 10^{-4}$ with Adam optimizer \cite{kingma2014adam}, no regularization method is used here.

\subsection{Main Results}
Table 1 contains the experimental results on 8 benchmark datasets from \cite{zhang2015character}. All results reported are averaged on five runs. From the results we can see that the use of meta-network brings improvements on 7 out of 8 benchmarks, with a largest performance gain of 1.7\%, while also achieves state-of-the-art results on 5 datasets.

\begin{figure*}[!h]
\centering
\includegraphics[width=\textwidth]{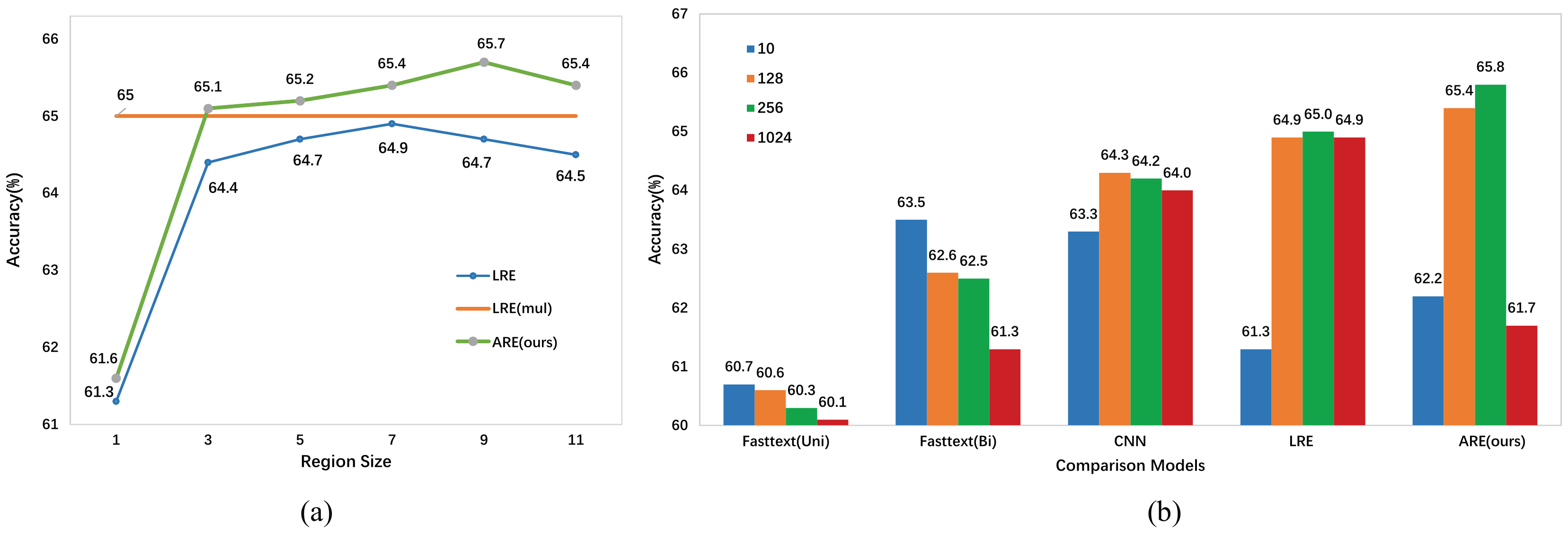}
\caption{Effect of embedding size and region size. For (a) the embedding size is fixed to 128, and LRE(mul) refers to an ensemble of region sizes of [3,5,7], for (b) the region size is fixed to 7. All competitor results come from \cite{qiao2018a}'s paper.}
\end{figure*}

\subsection{Comparison of the Number of Parameters}

One of the motivations of adopting meta-network to generate context unit is to reduce the large memory cost in \cite{qiao2018a} where over 80\% parameters are used as the look-up tensor to generate LCU, due to its storage of the whole vocabulary's context embedding. The counterpart in our method, ACU, however, greatly reduces the parameter space since the look-up tensor is replaced by a more compact meta-network which functions similarly.

If we denote $v, h, r, n$ to be the vocabulary size, word embedding size, region size and class number respectively ($r = 2c+1$), then the total number of parameters for \cite{qiao2018a} and our method can be calculated as follows:

\begin{small}
\begin{align}
    LRE :~~ & \underbrace{v \times h}_{Emb.}+ \underbrace{v\times h \times r}_{LCU} + \underbrace{h\times n + n}_{FC} \\
    ARE(Ours):~~ & \underbrace{v \! \times \! h}_{Emb.} + \underbrace{h \!\times\! (h \!\times\! r) \!\times\! r}_{ACU_{Conv}} + \underbrace{ (2\!\times\! \!h\times\! r) }_{ACU_{BN}} + \underbrace{h\!\times\! n\! +\! n}_{FC}
\end{align}
\end{small}

For text classification task, especially large-scale datasets such as Yahoo Answers, $v$ is several magnitudes' larger than other hyperparameters like $h, r, n$, and is the dominant factor of the total number of parameters.

A comparison of the number of parameters between \cite{qiao2018a} and ours can be found in Table 3. The first two columns demonstrate the total parameters size in the whole architecture, and the last two columns indicate the parameters size that is used to generate the context unit, corresponding to the embedding look-up tensor $\mathbf{U} \in R^{h \times (2c+1)\times v}$ and the meta-network respectively. From the table, we find that our total parameter space is only \textasciitilde{}26\% large as that in \cite{qiao2018a}, and if we only take the context unit generation part into account and leave out the rest, our ACU generation network reduced the LCU's parameters' number to less than 5\%. It is worth to mention that if we choose the region and embedding size to be 7 and 128 as that in \cite{qiao2018a}, the parameter space will be further reduced to nearly half of that in Table 3 and our method still outperforms \cite{qiao2018a}'s. Besides, our meta-network's parameter space is invariant to the vocabulary size, making it possible to be applied in large-scale datasets with a large vocabulary size.

\subsection{Effect of Embedding Size and Region Size}
Apart from the main classification results, we also experiment with different embedding and region sizes and compare with \cite{qiao2018a}'s method. The results from Figure 3 show that our method outperforms LRE and its ensemble of multiple region sizes with significant margins. Moreover, our method also performs better with different embedding sizes except for 1024, in which case the output space of the \emph{dynamically generated parameters} is too large for the meta-network to learn. Besides, we find that the optimal region and embedding size are different for ARE and LRE (9/256 and 7/128) which indicates our ARE requires a larger embedding size to contain more context information, and the meta-network is able to capture long-distance patterns, leading to a larger optimal region size.

\begin{figure*}[!h]
\centering
\includegraphics[width=\textwidth]{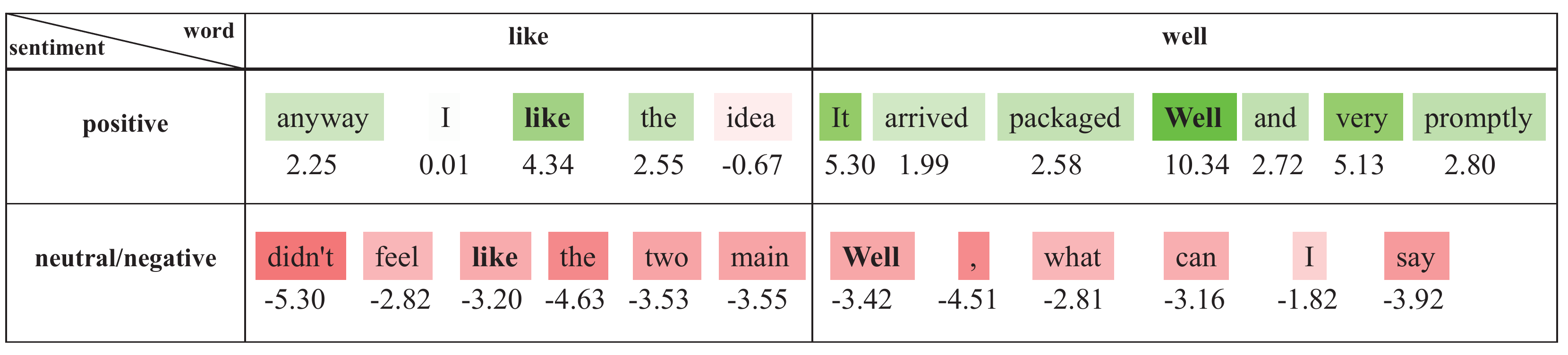}
\caption{Heatmaps of Samples from Amazon Review Polarity. Green denotes positive contribution and red denotes negative. It demonstrates that our model is able to distinguish different meanings of the same word under different contexts.}
\end{figure*}

\subsection{Visualization of Word Ambiguity Avoidance}
In order to illustrate that our method is able to distinguish context-sensitive words, we select the two words \emph{like} and \emph{well} with different meanings and visualize their contributions to the sentiment analysis on Amazon Review Polarity dataset. We adopt and modify the \emph{First-Derivative Saliency} strategy proposed in \cite{Li2016VisualizingAU} for visualization. The results are shown in Figure 4.

We find that in the first row, the \emph{like} conveys the positive attitude in the sentence \emph{I like the idea}, its derivative is greater than other words, which means that it contributes most to the final prediction of positive, the word \emph{well} means \emph{of high standard} which also has positive effects on the classification result, thus corresponding to the highest value. In the second row, however, both words convey neutral or negative attitude, where \emph{like} is part of the phrase \emph{feel like} and \emph{well} is a spoken expression, and the visualizations show that these two words do not contribute as much as that in the first row to the final sentiment prediction. These visualizations also accord with our intuition.

\subsection{Choice of the Meta-network}

It is worth noting that the choice of meta-network can be any differentiable architecture that is trainable using gradient descent. It can be multi-layer perceptron, convolutional network, or recurrent network etc. In this paper we choose a one-layer convolutional neural network to be our meta-network for its superior performance and simple network structure. 

In order to further investigate the influence of the choice of meta-network, we also experiment with the following variants:
\begin{itemize}

\item SmallCNN: Instead of producing ACU of size $\mathbf{K}_{w_i} \in R^{h \times (2c+1)}$, the SmallCNN meta-network produces $\mathbf{K}_{w_i} \in R^{h \times 1}$, and uses the same set of filters for all the positions from $[-c:c]$. This meta-network contains less parameters than our proposed CNN meta-network.
\item FactoredCNN: The produced filters are $\mathbf{K}_{w_i} \in R^{u \times 1}$ where $u$ is relatively small compared to $h$. Then the filters are transformed to $K' \in R^{h \times (2c+1)}$ by multiplying a matrix $\mathbf{P} \in R^{u \times(h\times (2c+1))}$. This model can be interpreted as factorizing the filters $\mathbf{K'}$ into the multiplication of two matrices: $\mathbf{K'} = \mathbf{KP}$, where $\mathbf{K}$ is generated by the meta-network, and $\mathbf{P}$ is updated by gradient descent. This meta-network further reduces the number of parameters. In the experiment, we set $u=32$. 
\item LSTM: We use an LSTM to generate the filters $\mathbf{K}$ where the hidden unit of LSTM is of size $h\times(2c+1)$.
\item GRU: We use a GRU to generate filters $\mathbf{K}$ where the hidden unit of LSTM is of size $h\times(2c+1)$.
\item Ensemble (CNN+LSTM): We generate two set of filters $\mathbf{K}_{CNN}$, $\mathbf{K}_{LSTM}$, and use the element-wise product $\mathbf{K}_{CNN} \odot \mathbf{K}_{LSTM}$ as the context unit.

\end{itemize}

We report the experimental results on AG's News, DBPedia and Yahoo Answers dataset due to page limit. From the result we find that the recurrent structure is not so suitable for generating adaptive filters, where the CNN variants with smaller parameter space yield comparable performances.

\begin{table}

  \label{sample-table-h}
  \centering
  \begin{tabular}{llll}
    \hline
    
    Hypernet     & AG  & DBP & Yahoo.A.  \\
    \hline
    \hline
    CNN & \textbf{93.1} & \textbf{99.1} & 74.9   \\
    SmallCNN     & 93.0  & 98.9 & 74.9\\
    FactoredCNN     & 92.7 & 98.9 & 74.3\\
    LSTM & 92.5 & 98.6  & 73.5  \\
    GRU     & 92.5 & 98.6 & 73.5\\
    Ensemble (CNN+LSTM)  & \textbf{93.1}  & \textbf{99.1} & \textbf{75.1}\\
    \hline
  \end{tabular}
\caption{Impact of the Choice of the Meta-network}
\end{table}

\section{Conclusion and Future Work}
In this paper, we propose a novel region embedding method called \emph{Adaptive Region Embedding} using the meta-network structure which is able to adaptively capture regional compositionality with a more compact parameter space. We also discuss the internal relationships between these methods under the \emph{Generalized Text Filtering} framework, where our method corresponds to the \emph{instance-level filtering} which is more flexible. By experimenting on benchmark text classification datasets, we are able to gain higher classification performances with a small parameter space,  while also avoiding word ambiguity. In future, we aim to design more efficient structures of meta-network and combine techniques such as attention mechanism into our model.

\section{Acknowledgments}
The work is supported by National Basic Research Program of China (2015CB352300) and National Natural Science Foundation of China (61571269). We also thank reviewers for their constructive comments.

\bibliographystyle{aaai}
\bibliography{aaai}
\end{document}